\documentclass[runningheads]{llncs}

\usepackage{eccv}

\usepackage{eccvabbrv}

\usepackage{graphicx}
\usepackage{booktabs}

\usepackage[accsupp]{axessibility}  

\usepackage{hyperref}

\usepackage{orcidlink}

\usepackage{mathrsfs}
\usepackage{color}
\usepackage{booktabs}
\usepackage{multirow}
\usepackage{threeparttable}
\usepackage{color, colortbl}
\usepackage{changes}
\usepackage{amssymb}

\usepackage{graphicx}
\usepackage{amsmath}
\usepackage{pifont}
\usepackage{xcolor}

\newcommand{\highest}[1]{\color{red}{\textbf{{#1}}}}
\newcommand{\second}[1]{\color{blue}{\underline{{#1}}}}

\begin{document}

\title{Improving Video Segmentation via Dynamic Anchor Queries}
\titlerunning{DVIS-DAQ}

\author{
Yikang Zhou\inst{1~\star}\orcidlink{0000-0001-8326-5925} \and
Tao Zhang\inst{1,2}~\thanks{The first two authors contribute equally. This work was performed when Tao Zhang was an Intern at Skywork AI. \textsuperscript{$\dagger$} Project Leader. Corresponding authors: Shunping Ji and Xiangtai Li.}\orcidlink{0000-0001-7390-2409}  \and
Shunping Ji\inst{1}\orcidlink{0000-0002-3088-1481} \and \\
Shuicheng Yan\inst{2}\orcidlink{0000-0001-8906-3777} \and
Xiangtai Li\inst{2}\orcidlink{0000-0002-0550-8247} \textsuperscript{$\dagger$}
}

\authorrunning{Y. Zhou and T. Zhang et al.}
\institute{\small Wuhan University \and Skywork AI \\
\email{\small zhang\_tao@whu.edu.cn, zhouyik@whu.edu.cn, xiangtai94@gmail.com} \\
\small Project page: \url{https://zhang-tao-whu.github.io/projects/DVIS_DAQ} \\ }

\maketitle

\begin{abstract}
  Modern video segmentation methods adopt object queries to perform inter-frame association and demonstrate satisfactory performance in tracking continuously appearing objects despite large-scale motion and transient occlusion.
  However, they all underperform on newly emerging and disappearing objects that are common in the real world because they attempt to model object emergence and disappearance through feature transitions between background and foreground queries that have significant feature gaps. We introduce Dynamic Anchor Queries (DAQ) to shorten the transition gap between the anchor and target queries by dynamically generating anchor queries based on the features of potential candidates. 
  %
  %
  Furthermore, we introduce a query-level object Emergence and Disappearance Simulation (EDS) strategy, which unleashes DAQ's potential without any additional cost.
  Finally, we combine our proposed DAQ and EDS with the previous method, DVIS, to obtain DVIS-DAQ.
  Extensive experiments demonstrate that DVIS-DAQ achieves a new state-of-the-art (SOTA) performance on five mainstream video segmentation benchmarks. 
  %
  \keywords{Video segmentation \and Dynamic anchor design \and Universal segmentation}
\end{abstract}
\section{Introduction}
\label{sec:intro}

\begin{figure}[t]
    \centering
    \includegraphics[width=0.98\linewidth]{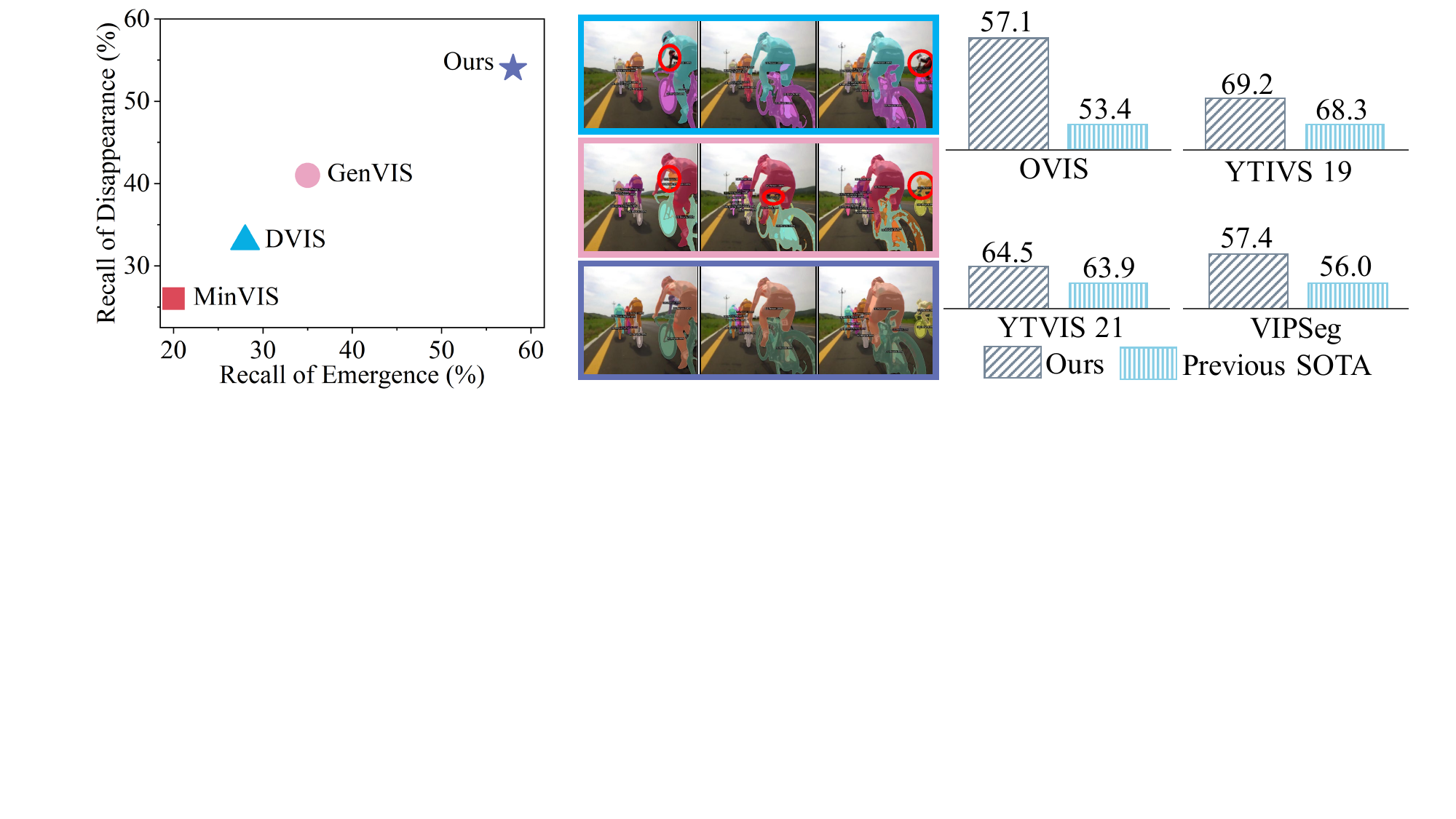}
    \caption{\textbf{The preliminary experiments and our results.} We analyze current query-based video segmentation methods' capabilities in handling objects' emergence and disappearance within videos. The scatter plot on the left illustrates the recall ratios for newly emerging and disappearing video objects on a subset of the BDD~\cite{bdd100k} dataset. The visual comparison is presented in the middle. The top two rows output from DVIS~\cite{zhang2023dvis} and GenVIS~\cite{heo2023generalized}, respectively. And the bottom row output from our method. The poor performance of previous methods on new emergence detection and disappearance filtering is highlighted by a red circle. On the right side, our method outperforms the current SOTA on mainstream video segmentation datasets.}
    \label{fig:oracle-exp}
\end{figure}

Video segmentation aims at simultaneously identifying, segmenting, and tracking all objects of interest within a video \cite{yang2019video, kim2020video, qi2022occluded, yang2023pvsg}, which has many applications, including autonomous driving, video editing, content analysis, and video understanding~\cite{patrick2021keeping,gberta_2021_ICML}.
Modern video segmentation methods~\cite{heo2023generalized, li2023tcovis, hannan2023gratt, zhang2023dvis, ying2023ctvis, zhang2023dvis++, li2023tube, huang2022minvis, wu2022defense} utilize object queries to perform cross-frame association and achieve remarkable performance. 
Despite large-scale motion and transient occlusion, these methods yield satisfactory results in continuously appearing objects.

However, these query-based methods exhibit a notable limitation: they tend to underperform when dealing with \textit{newly emerging} or 
\textit{disappearing objects}, even with recent state-of-the-art (SOTA) methods~\cite{zhang2023dvis,heo2023generalized}.
As shown in the left part of Fig.~\ref{fig:oracle-exp}, these methods only achieve less than 45\% recall ratio on emerging and disappearing objects. 
As illustrated in the middle part of Fig.~\ref{fig:oracle-exp}, they often fail to detect newly emerging objects. 
And they may incorrectly track a different, similar object when the original object disappears. 
%

\begin{figure}[t]
    \centering
	\includegraphics[width=0.98\linewidth]{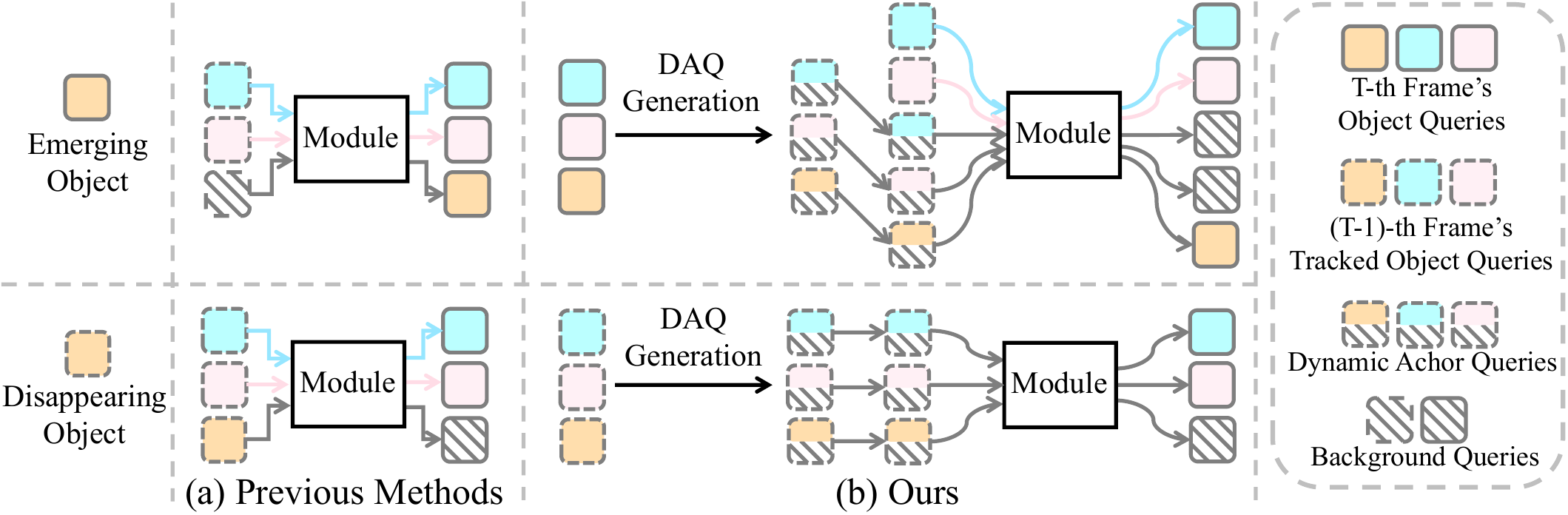}
    \caption{\textbf{The process of handling newly emerging and disappearing objects in different methods.} Unlike previous approaches that treat appearing, disappearing, and tracked objects equally, our method dynamically generates anchor queries for emergence and disappearance based on candidate objects' features to effectively shorten the transition gap.}
    \label{fig:idea}
\end{figure}

A natural question arises: "What causes the huge performance gap between continuously appearing objects and emerging/disappearing objects?"
We argue the answer lies in the unreasonable anchor queries used in current methods. 
As shown in Fig.~\ref{fig:idea}, SOTA query-based methods~\cite{heo2023generalized,zhang2023dvis,zhang2023dvis++,huang2022minvis} use the background query as the anchor query and model the object's emergence and disappearance as a feature transition between the background and foreground query. 
However, a significant transition gap between background and foreground queries poses considerable challenges during training.
To illustrate, consider emergence as an example.
It is difficult for the model to learn how to transform a background anchor query into a foreground query without any semantic similarity.
Consequently, the model may fail to transform and retain the background anchor query, resulting in missing newly emerging objects.
Similarly, for disappearing objects, the model may prefer to incorrectly transform their query into another object's query with more semantic similarity.
%

In this paper, we address this challenge by introducing \textbf{D}ynamic \textbf{A}nchor \textbf{Q}ueries~(\textbf{DAQ}) mechanism to detect the emergence and disappearance of objects. 
As illustrated on the right side of Fig.~\ref{fig:idea}, we regard all objects segmented in the current frame as potential candidates for newly emerging objects. 
We dynamically generate emergence anchor queries based on the features of these candidates to reduce the transition gap significantly. 
In particular, we designate all tracked objects as candidates for disappearing objects and generate disappearance anchor queries dynamically using their features. 
These emergence and disappearance anchor queries and the representations of tracked objects are collectively inputted into the tracker to identify the objects of interest in the current frame. 
Objects identified by the emergence anchor queries are classified as newly appeared, while those pinpointed by the disappearance anchor queries are recognized as having disappeared, prompting their removal. 
The remaining queries are responsible for tracking objects present across both past and current frames.

The proposed DAQ mechanism enhances the network's ability to handle the emergence and disappearance of objects. 
However, emergence and disappearance events are rare in existing video segmentation datasets. 
To further leverage the capabilities of DAQ, we propose a training time simulation strategy for the emergence and disappearance of objects, termed \textbf{E}mergence and \textbf{D}isappearance \textbf{S}imulation (\textbf{EDS}). 
This strategy simulates emergence and disappearance events by removing high-level object queries. 
We mimic object emergence by randomly eliminating some object queries propagated from the previous frame. 
Likewise, we simulate object disappearance by removing some of the current frame's object queries. 
Our simple yet effective simulation strategy generates numerous instances of object emergence and disappearance during training, thereby enabling comprehensive training of the DAQ mechanism and maximizing its potential.

To verify the effectiveness of our proposed DAQ mechanism and EDS strategy, we integrate them into the current SOTA method, DVIS~\cite{zhang2023dvis}, to construct the DVIS-DAQ. 
As shown in the right part of Fig.~\ref{fig:oracle-exp}, our proposed DVIS-DAQ achieves new SOTA performance on five mainstream video segmentation benchmarks and significantly outperforms previous SOTA methods.

In summary, our main contributions are as follows:
\begin{enumerate}
\item We have found that current query-based methods fall short in managing objects' emergence and disappearance due to the considerable transition gap between foreground and background queries. We propose a Dynamic Anchor Queries (DAQ) mechanism to address this challenge. 
The DAQ mechanism dynamically generates anchor queries for objects' emergence and disappearance, thereby shortening the gap and effectively overcoming the challenge.

\item To further unleash the potential of the DAQ mechanism, we introduce the Emergence and Disappearance Simulation (EDS) strategy. 
This straightforward and efficient approach amplifies the number of emergence and disappearance cases, thereby ensuring comprehensive training of the DAQ.

\item We enhance the SOTA video segmentation method DVIS by integrating our proposed DAQ mechanism and EDS strategy, resulting in DVIS-DAQ.
With extensive experiments on five video segmentation benchmarks, DVIS-DAQ achieves new SOTA performance on all benchmarks. 
We further present detailed ablation studies to verify the effectiveness of each component.

\end{enumerate}

\section{Related Work}
\label{sec:related_work}

\noindent
\textbf{Video Instance Segmentation (VIS).} Earlier works~\cite{yang2019video,lin2020video_vae,yang2021crossover} extend image instance segmentation methods by adding tracking heads and learning the feature association. With the rise of vision transformer~\cite{carion2020end,wang2021end,dosovitskiy2020image}, current SOTA video instance methods~\cite{wang2021end,heo2022vita,cheng2021mask2former,wu2022seqformer,huang2022minvis, wu2022defense, ying2023ctvis, li2023tube, meinhardt2023novis,li2023tcovis,hannan2023gratt} adopt query-based designs. In particular, Video K-Net~\cite{li2022video} and IDOL~\cite{wu2022defense} directly learn the query association embeddings via contrastive learning. After that, association-based methods \cite{yang2019video, huang2022minvis, wu2022defense, ying2023ctvis, li2023tube, meinhardt2023novis}, accomplish the tracking of video targets between adjacent frames or adjacent video clips through object matching. Meanwhile, several models~\cite{heo2022vita,cheng2021mask2former,wu2022seqformer} directly predict 3D volumes of video instances across times in a semi-online manner. On the other hand, propagation-based methods \cite{heo2023generalized,li2023tcovis,hannan2023gratt,zhang2023dvis} delegate the tracking problem to the transformer decoder. They take the object queries outputted from the previous frame or video clip as input and iteratively optimize them to segment the objects in the current frame. 
Inspired by the design of the dual encoder in the low-level area~\cite{tsai2023dual} and multi-task learning~\cite{li2022panoptic,xu2022fashionformer}, several works have also decoupled the tracking decoder and segmentation decoder.
Note that these methods can also be applied for video panoptic segmentation (VPS) and video semantic segmentation (VSS) tasks~\cite{qiao2021vip,miao2021vspw,miao2022large,weber2021step,kim2020video,Zhu2016DeepFF} by adding queries for segmenting the stuff. However, both poorly handle challenges such as new-emerging objects and target disappearance. Our approach improves the model's ability to address these challenges by employing dynamic anchor queries to manage the emergence and disappearance of objects.


\noindent
\textbf{Universal Video Segmentation.} Current efforts in the field of video segmentation directly draw upon the design of universal image segmentation models~\cite{weng2023mask,zhang2021k,cheng2021mask2former,cheng2021per,yan2023universal,cheng2023tracking,wang2023cut} to develop universal video segmentation models \cite{li2022video,kim2022tubeformer,li2023tube,athar2023tarvis,zhang2023dvis,li2023transformer,wu2023open}, achieving comparable or even superior performance to specialized video segmentation methods. Video K-Net~\cite{li2022video} and TubeFormer~\cite{kim2022tubeformer} are the first to unify all three video segmentation tasks. TarVIS~\cite{athar2023tarvis} extends the concept of mask classification from image segmentation to video segmentation, accomplishing universal video segmentation via one shared model. DVIS~\cite{zhang2023dvis} receives frame object query outputs from Mask2Former~\cite{cheng2021mask2former} for universal video segmentation. Recently, several works~\cite{li2024omg,xu2024rapsam} unified both image and video segmentation in one model. However, these methods are still unaware of object emergence and disappearance problems, which are significant for long videos.

\noindent
\textbf{Video Object Tracking.} Object tracking~\cite{zeng2021motr,seong2021hierarchical,pang2021simpletrack,zhang2022bytetrack,peize2021dance,unicorn,qdtrack_conf,MeMOTR,cheng2021stcn,Cae+17,bekuzarov2023xmem,zhou2022transvod}
is a crucial task in VIS and VPS~\cite{zhou2021slot,kim2020video,kim2020vps,bertasius2021classifying,li2024univs,he2023maxtron}, and many works adopt the tracking-by-detection paradigm~\cite{bewley2016simple,leal2016learning,xu2019spatial,zhu2018online,porzi2020learning,TraDeS,transtrack,meinhardt2021trackformer}. 
These methods divide the task into two sub-tasks, where an object detector first detects objects and then associates them using a tracking algorithm. 
Recent works~\cite{park2022per,MOSE,cliptrack,Athar_Mahadevan20ECCV} also perform clip-wise segmentation and tracking. However, the former~\cite{park2022per,MOSE} only focuses on single-object mask tracking, while the latter~\cite{cliptrack} considers global tracking via clip-level matching. 
%
\section{Method}
\label{sec:method}

\subsection{Preliminary}
\label{sec:p&m}


\noindent
\textbf{Query-based Video Segmentation Formulations.}
Modern mainstream video segmentation methods~\cite{zhang2023dvis,zhang2023dvis++,heo2023generalized,li2023tube,wu2022defense} rely on queries to represent objects and achieve cross-frame association of objects. Here, we summarize current query-based video segmentation methods with the following formulation:
\begin{equation}
    Q^{T}_{Seg} = \mathcal{S}(\mathcal{I}^{T}),
    \quad Q^{T} = \mathcal{A}(Q^{T-1},Q^{T}_{Seg}),
    \quad \mathcal{M}^{T}, \mathcal{C}^{T} = \mathcal{D}(Q^{T}).\\ 
\end{equation}
\label{eq: 1}First, a segmenter $\mathcal{S}$ extracts query representations of objects $Q^{T}_{Seg}$ from a single-frame image. Then, an association component $\mathcal{A}$ establishes connections between the current frame's object representations $Q^{T}_{Seg}$ and $Q^{T-1}$ to obtain associated object representations $Q^{T}$. Finally, the object representations $Q^{T}$ are decoded to obtain predictions of object categories $\mathcal{C}^{T}$ and segmentation masks $\mathcal{M}^{T}$ for the current frame. In this pipeline, the association component $\mathcal{A}$ is the most critical and challenging part. Most video segmentation methods focus on designing more efficient association modules.

The current mainstream association modules $\mathcal{A}$ can be divided into association-based~\cite{huang2022minvis,wu2022defense,li2023tube,ying2023ctvis} and propagation-based~\cite{zhang2023dvis,zhang2023dvis++,heo2023generalized} pipelines. They both model object emergence as a transition from background queries to foreground queries and model disappearance as a transition from foreground queries to background queries. Specifically, we can split $Q$ ($Q = \{Q_{Bg}, Q_{Fg}\}$) into $Q_{Fg}$ and $Q_{Bg}$, and the tracking of continuously appearing objects, newly emerging objects, and disappearing objects can be described as follows:
\begin{equation}
    Cont: Q_{Fg}^{T-1} \rightarrow Q_{Fg}^{T}, \quad Emg: Q_{Bg}^{T-1} \rightarrow Q_{Fg}^{T}, \quad Dis: Q_{Fg}^{T-1} \rightarrow Q_{Bg}^{T}\\ 
    \label{eq: model}
\end{equation}

\noindent
\textbf{Limitations of the Query-based Methods.}
We find that current query-based methods~\cite{zhang2023dvis,heo2023generalized,huang2022minvis} perform poorly on objects that emerge and disappear in the middle of a video, as shown in Fig.~\ref{fig:oracle-exp}. For emerging objects, SOTA methods often miss detecting them. For disappearing objects, SOTA methods frequently fail to recognize that the object has disappeared and may incorrectly segment another object. We quantitatively analyzed the recall ratios of emerging and disappearing objects in preliminary experiments using several SOTA methods. We found that the recall ratios of emerging and disappearing objects are very low (below 45\%) for these methods, as shown in Fig.~\ref{fig:oracle-exp}.

\noindent
\textbf{Thorough Analysis and Our Motivation.}
Query-based methods have demonstrated satisfactory performance on consistently appearing objects (exceeding 60 AP on the YouTube-VIS~\cite{yang2019video}), but they exhibit poor performance on newly emerging and disappearing objects. As depicted in Fig.~\ref{fig:idea} and Eq.~\ref{eq: model}, in current query-based methods, there is a significant disparity in the modeling formulation between newly emerging or disappearing objects and consistently appearing objects. The former poses more challenges because of the significant feature gap in the transition between foreground and background queries. In this work, we dynamically generate anchor queries for newly emerging and disappearing objects, resulting in a lower feature transition gap than the background query. We term the background query used in previous methods as Static Anchor Query (SAQ), the query of continuously appearing objects as Continuously Tracked Query (CTQ), and our proposed dynamically generated anchor query as Dynamic Anchor Query (DAQ). We use our proposed DAQ to replace the SAQ to address the challenge above. The details of the DAQ will be introduced in Sec.~\ref{sec:daq}.

Furthermore, we observed that only less than 20\% of samples contain object emergence and disappearance in the existing datasets~\cite{yang2019video,qi2022occluded,miao2022large}, including the most challenging OVIS~\cite{qi2022occluded} dataset. Therefore, to fully unleash the potential of our proposed DAQ, we introduce a simple yet effective query-level object emergence and disappearance simulation strategy, which will be detailed in Sec.~\ref{sec:simulation}.

\subsection{Dynamic Anchor Queries}
\label{sec:daq}
The key idea of DAQ is to dynamically generate anchor queries using the features of candidate objects that may emerge or disappear, thereby reducing the gap between anchor queries and the target queries (the queries of actual newly emerging or disappearing objects). DAQ allows the network to easily transform anchor queries to target queries and effectively handle object emergence and disappearance.

In this subsection, we introduce how to design and utilize Dynamic Anchor Queries (DAQ) to address the challenges mentioned above. First, we explain how the DAQ for newly emerging and disappearing objects are generated. Then, we explore how to make minor adjustments to the current query-based tracker to accommodate our proposed DAQ.

\begin{figure}[t]
    \centering
    \includegraphics[width=0.98\linewidth]{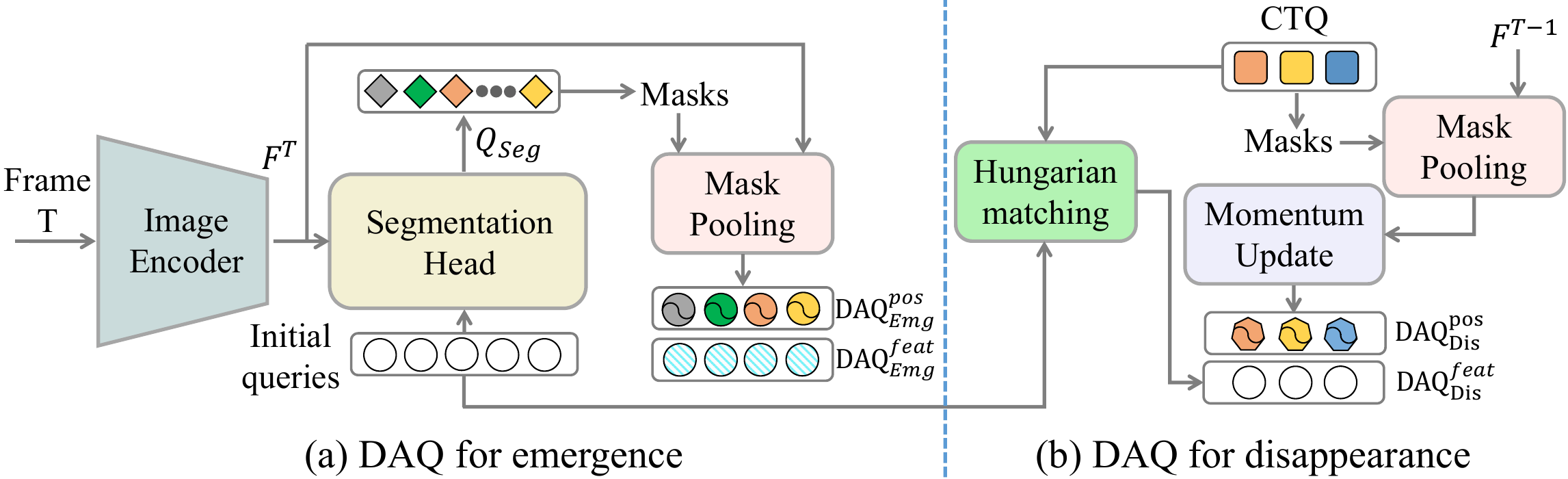}
    \caption{\textbf{Generation of dynamic anchor queries.} $F^{T}$ represents the image feature of the T$^{th}$ frame. Symbols \emph{Emg} and \emph{Dis} denote emergence and disappearance. Symbols \emph{feat} and \emph{pos} indicate feature and positional embedding, respectively.}
    \label{fig:daq}
\end{figure}

\noindent
\textbf{Dynamic Anchor Queries for Emergence.}
When processing the T$^{th}$ frame of a video, we consider all objects appearing in the current frame as candidates for newly emerging objects. The dynamic anchor queries for emergence DAQ$_{Emg}$ are naturally generated based on the features of these candidates. As shown in Fig.~\ref{fig:daq} (a), DAQ$_{Emg}$ consists of two parts: query feature DAQ$_{Emg}^{feat}$ and query positional embedding DAQ$_{Emg}^{pos}$. The query feature DAQ$_{Emg}^{feat}$ is a learnable embedding shared by all dynamic anchor queries. The query positional embedding DAQ$_{Emg}^{pos}$ is obtained using the appearance features $F_{mask}^{T}$ of the candidates through simple Mask-Pooling:
\begin{equation}
    F_{mask}^{T} = \mathrm{MLP}(\mathrm{MaskPooling}(F^{T}, \mathcal{M}^{T})), \\
\label{eq: 4}
\end{equation} 
where $F^{T}$ represents the image feature of the $T-th$ frame, and $\mathcal{M}^{T}$ represents the segmentation masks of the candidates outputted by the segmenter.

\noindent
\textbf{Dynamic Anchor Queries for Disappearance.}
When processing the T$^{th}$ frame, we consider all currently tracked objects as candidates for disappearing objects. Similar to the dynamic anchor queries for emergence, the DAQ$_{Dis}$ for disappearing objects are naturally generated based on the features of these candidates. As shown in Fig.~\ref{fig:daq} (b), the DAQ$_{Dis}$ also consists of query feature DAQ$_{Dis}^{feat}$ and query positional embedding DAQ$_{Dis}^{pos}$. We use the candidates' momentum-weighted appearance features $\overline{F}_{mask}^{T}$ as the query positional embedding DAQ$_{Dis}^{pos}$. The momentum-weighted function follows CTVIS~\cite{ying2023ctvis}:
\begin{equation}
    \overline{F}_{mask}^{T} = (1-\beta)\overline{F}_{mask}^{T-1} + \beta F_{mask}^{T} \\
    \label{eq: 5}
\end{equation}
\begin{equation}
\beta = \max\left(0, \frac{1}{T} \sum^{T-1}_{t=1} cosine(F_{mask}^{t}, F_{mask}^{T})\right) \\
\label{eq: 6}
\end{equation}
Since the disappearance of an object needs to be modeled as a transition from an anchor query to a background query, to reduce the transition gap, we use initial queries of segmenter as the query feature DAQ$_{Dis}^{feat}$. Specifically, for each query feature, we match the closest one from the initial queries of the segmenter based on cosine similarity.

\noindent
\textbf{Continuously Tracked Queries.} We also assign query positional embedding CTQ$^{pos}$ for continuously tracked queries (CTQ) to maintain the format consistency with dynamic anchor queries. We use the momentum-weighted appearance feature of the tracked objects as CTQ$^{pos}$, the same process as Eq.~\ref{eq: 4}, \ref{eq: 5}, and \ref{eq: 6}.

\begin{figure}[t]
    \centering
    \includegraphics[width=0.98\linewidth]{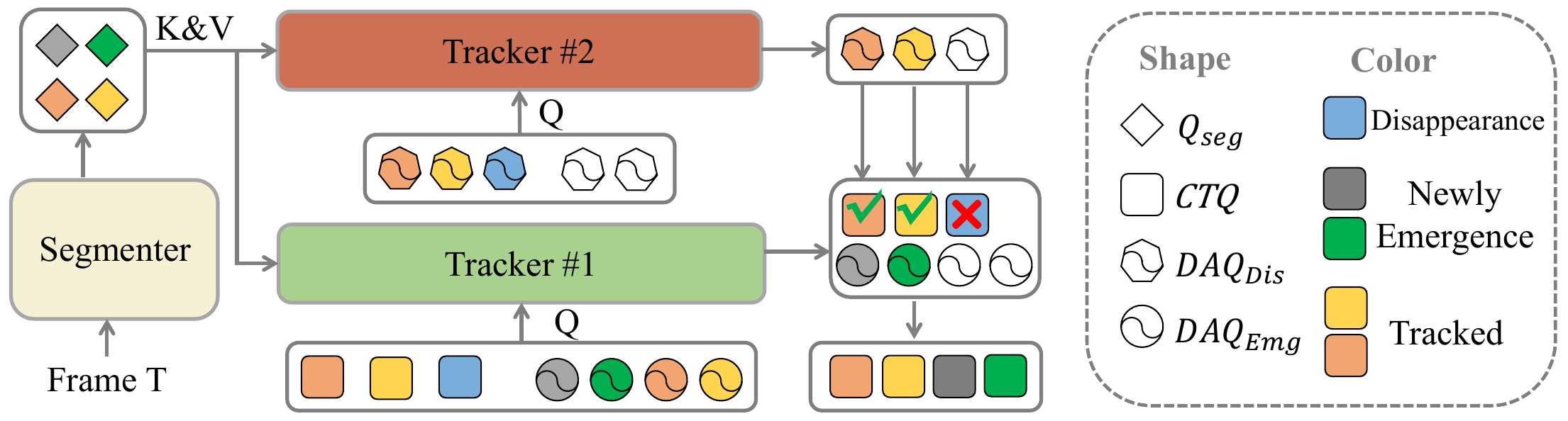}
    \caption{\textbf{Tracker with dynamic anchor queries.} $Q_{seg}$ represents the query output of segmenter. CTQ and DAQ stand for continuously tracked and dynamic anchor queries, respectively. Symbols \emph{Emg} and \emph{Dis} denote emergence and disappearance.}
    \label{fig:daq-tracker}
\end{figure}

\noindent
\textbf{Incorporating Dynamic Anchor Queries into a Tracker.} As depicted in Fig.~\ref{fig:daq-tracker}, we can seamlessly integrate DAQ into mainstream propagation-based trackers~\cite{zhang2023dvis,heo2023generalized}, requiring no alterations to the tracker's network architecture. We instantiate two trackers: Tracker 1, which takes CTQ and DAQ$_{Emg}$ as inputs and is responsible for tracking continuously appearing and newly emerging objects, and Tracker 2, which receives DAQ$_{Dis}$ along with several learnable background embeddings as inputs and identifies disappearing objects in the current frame. It's important to note that the SoftMax function in Tracker 2 operates along the Q-dimension to prevent multiple dynamic anchor queries from segmenting the same object. Through this pipeline, we can effectively and reasonably manage and model the emergence and disappearance of objects. Despite introducing an additional tracker, it's worth highlighting that the computational overhead is minimal, constituting less than 2\% of the overall computational cost~\cite{zhang2023dvis}, as processing occurs solely at the query level rather than on dense image features.

\subsection{Emergence and Disappearance Simulation}
\label{sec:simulation}

\begin{figure}
    \centering
    \includegraphics[width=0.98\linewidth]{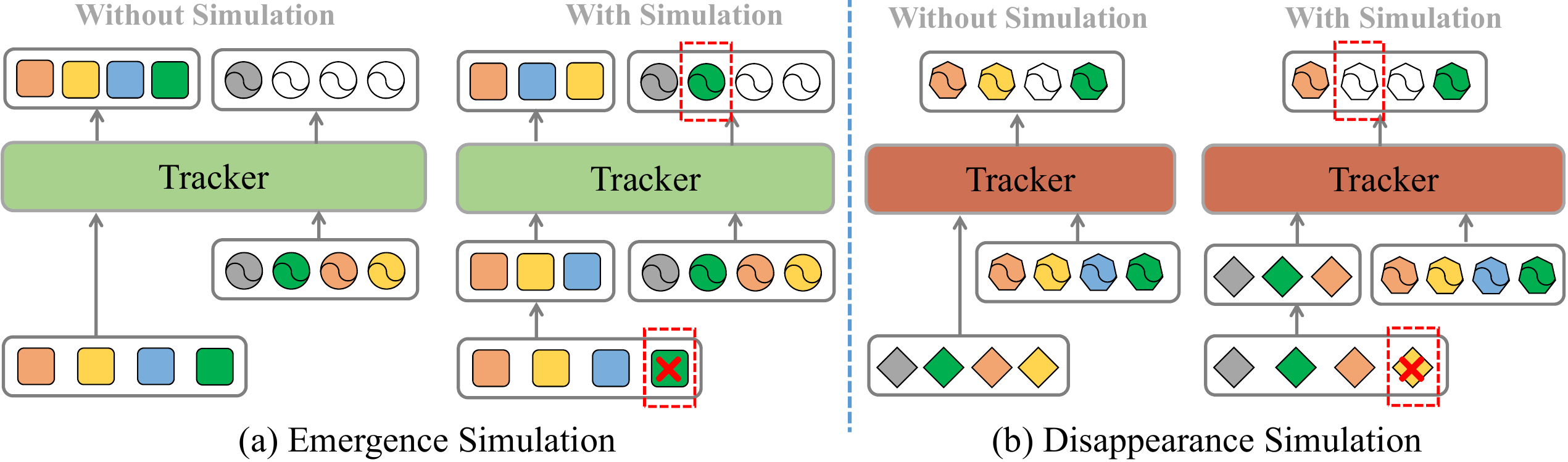}
    \caption{\textbf{The pipelines of emergence and disappearance simulation.} CTQ, DAQ$_{Emg}$, DAQ$_{Dis}$, and $Q_{Seg}$ are represented by rectangles, circles, diamonds, and heptagons, respectively. White indicates the background query transformed from the anchor query. The red rectangles highlight the difference between with and without simulation.}
    \label{fig:simulation}
\end{figure}

As shown in Fig.~\ref{fig:simulation}, when using DAQ, it is possible to simulate object emergence and disappearance separately by removing parts of CTQ and $Q_{Seg}$, respectively. We term this simple yet efficient simulation strategy as Emergence and Disappearance Simulation (EDS), which includes Emergence Simulation (ES) and Disappearance Simulation (DS). EDS can simulate numerous object emergence and disappearance cases during training without introducing additional costs, allowing the network to be fully trained and unleashing the potential of DAQ.

\noindent
\textbf{New Emergence Simulation.}
As shown in Fig.~\ref{fig:simulation} (a), taking the green object as an example, we can simulate it as an emerging object by removing its corresponding query from CTQ. At this point, this green object is not present in CTQ but exists in the current frame, consistent with the state of a real newly emerged object colored in gray. Then, the Tracker needs to transition the corresponding anchor query from DAQ$_{Emg}$ to a foreground query and predict this green object's segmentation mask and class.

\noindent
\textbf{Disappearance Simulation.}
As shown in Fig.~\ref{fig:simulation} (b), taking the yellow object as an example, by removing its corresponding query from $Q_{Seg}$, we can simulate this object as a disappearing object because the Tracker only detects features from $Q_{Seg}$ through cross-attention to obtain image information. At this point, this blue object is not present in the current frame but exists in the set of tracked objects, consistent with the state of a real disappearing blue object. The Tracker needs to transition the corresponding anchor query to a background query and determine the disappearance of this object.

\subsection{Overall Architecture}
\label{sec:dvis-daq}

\begin{figure}[t]
    \centering
    \includegraphics[width=0.98\linewidth]{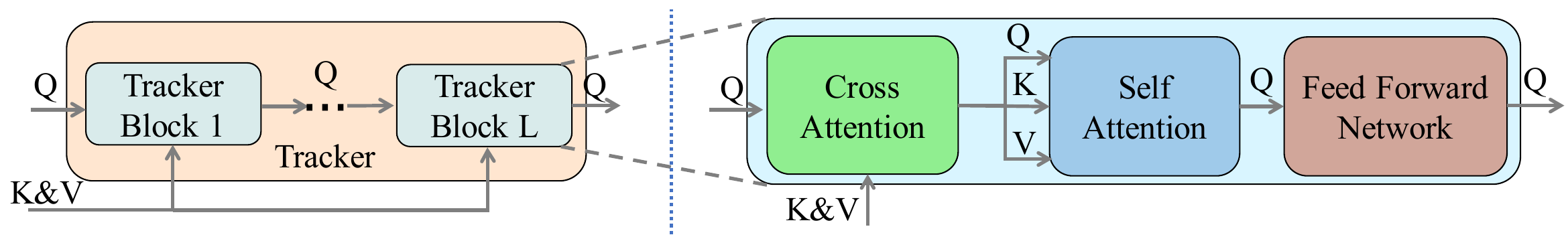}
    \caption{\textbf{The architecture of DVIS-DAQ tracker.} }
    \label{fig:overall_arc}
\end{figure}

\noindent
\textbf{Architecture.} Sec.~\ref{sec:daq} details how DAQ is integrated into the current query-based tracker without any requirement for architecture modification. We apply the SOTA video segmentation method DVIS~\cite{zhang2023dvis} as our baseline and replace the Referring Cross-Attention~\cite{zhang2023dvis,zhang2023dvis++} in its tracker with standard Cross-Attention to achieve a more straightforward and generalizable structure shown in Fig.~\ref{fig:overall_arc}. We then integrated DAQ into this simplified tracker, resulting in our architecture, DVIS-DAQ.

\noindent
\textbf{Objective Function.} Our objective function differs from DVIS~\cite{zhang2023dvis}. We maintain the historical matching relationship between the predictions of CTQ and ground truth for continuously tracked objects and remove the matched items from the ground truth. For newly emerging and disappearing objects, since DAQ is dynamically generated based on the features of candidates (refer to Sec.~\ref{sec:daq}), we assign the remaining ground truth to DAQ based on the matching relationship between candidates and the remaining ground truth. Following this process, we obtain the matching relationship $\{(i, \sigma(i)) | i\in [1, M]\}$ between ground truth and predictions, M is the number of ground truth. The not-matched predictions are assigned with the background class and an empty segmentation mask. Then, the loss between matched prediction-label pairs $(\mathcal{P}, \mathcal{G})$ can be calculated following the Mask2Former~\cite{cheng2022masked}:
\begin{equation}
    \mathcal{L}(\mathcal{P}, \mathcal{G}) = \lambda_{cls}\mathcal{L}_{cls}(\mathcal{P}, \mathcal{G}) + \lambda_{dice}\mathcal{L}_{dice}(\mathcal{P}, \mathcal{G}) + \lambda_{bce}\mathcal{L}_{bce}(\mathcal{P}, \mathcal{G})
\end{equation}
Here, $\mathcal{L}_{cls}$ represents the classification loss term, implemented using Cross Entropy Loss. $\mathcal{L}_{dice}$ and $\mathcal{L}_{bce}$ denote the segmentation loss terms, implemented respectively using Dice loss and Binary Cross Entropy loss.
\begin{table*}[t]
\centering
\caption{Results on the val set of OVIS and YouTube-VIS 2019 \& 2021. $\dag$ denotes with the offline refiner proposed by~\cite{zhang2023dvis}. The VIT-L~\cite{dosovitskiy2020image} is pre-trained by DINOv2~\cite{oquab2023dinov2} and uses VIT-Adapter~\cite{chen2022vision} to obtain multiscale features.}
\label{tab:ovis&yt19&yt21}
\begin{tabular}{l l | c c c | c c c | c c c}
\toprule[1.5pt]
    \multirow{2}{*}{Method} & \multirow{2}{*}{Backbone} & \multicolumn{3}{c|}{OVIS} & \multicolumn{3}{c|}{YT-VIS 2019} & \multicolumn{3}{c}{YT-VIS 2021} \\
    ~ & ~ & AP &  AP$_{\rm 75}$ & AR$_{\rm 10}$ & AP &  AP$_{\rm 75}$ & AR$_{\rm 10}$ & AP &  AP$_{\rm 75}$ & AR$_{\rm 10}$\\
    \midrule[1pt]
    MinVIS~\cite{huang2022minvis} & R50 & 25.0 & 24.0 & 29.7 & 47.4 & 52.1 & 55.7 & 44.2 & 48.1 & 51.7\\
    CTVIS~\cite{ying2023ctvis} & R50 & 35.5 & 34.9 & 41.9 & 55.1 & 59.1 & 63.2 & \second{50.1} & \second{54.7} & \highest{59.5}\\
    GenVIS~\cite{heo2023generalized} & R50 & 35.8 & 36.2 & 39.6 & 50.0 & 54.6 & 59.7 & 47.1 & 51.5 & 54.7\\
    DVIS~\cite{zhang2023dvis} & R50 & 30.2 & 30.5 & 37.3 & 51.2 & 57.1 & 59.3 & 46.4 & 49.6 & 53.5\\
    DVIS\dag~\cite{zhang2023dvis} & R50 & 33.8 & 33.5 & 39.5 & 52.3 & 58.2 & 60.4 & 47.4 & 51.6 & 55.2\\
    DVIS++~\cite{zhang2023dvis++} & R50 & 37.2 & 37.3 & 42.9 & \highest{55.5} & \second{60.1} & \second{62.6} & 50.0 & 54.5 & \second{58.4}\\
    DVIS++\dag~\cite{zhang2023dvis++} & R50 & \second{41.2} & \second{40.9} & \second{47.3} & - & - & - & - & - & -\\
    \rowcolor{gray!35}
    DVIS-DAQ & R50 & 38.7 & 37.6 & 45.2 & \second{55.2} & \highest{61.9} & \highest{63.7} & \highest{50.4} & \highest{55.0} & 57.6 \\
    \rowcolor{gray!35}
    DVIS-DAQ\dag & R50 & \highest{43.5} & \highest{43.8} & \highest{49.1} & - & - & - & - & - & -\\
    \hline
    MinVIS~\cite{huang2022minvis} & Swin-L & 39.4 & 41.3 & 43.3 & 61.6 & 68.6 & 66.6 & 55.3 & 62.0 & 60.8\\
    CTVIS~\cite{ying2023ctvis} & Swin-L & 46.9 & 47.5 & 52.1 & 65.6 & 72.2 & 70.4 & 61.2 & 68.8 & 65.8\\
    GenVIS~\cite{heo2023generalized} & Swin-L & 45.2 & 48.4 & 48.6 & 64.0 & 68.3 & 69.4 & 59.6 & 65.8 & 65.0\\
    DVIS~\cite{zhang2023dvis} & Swin-L & 45.9 & 48.3 & 51.5 & 63.9 & 70.4 & 69.0 & 58.7 & 66.6 & 64.6 \\
    DVIS\dag~\cite{zhang2023dvis} & Swin-L & 48.6 & 50.5 & 53.8 & 64.9 & 72.7 & 70.3 & 60.1 & 68.4 & 65.7 \\
    DVIS++~\cite{zhang2023dvis++} & VIT-L & 49.6 & 55.0 & 54.6 & 67.7 & 75.3 & \second{73.7} & 62.3 & 70.2 & 68.0\\
    DVIS++\dag~\cite{zhang2023dvis++} & VIT-L & 53.4 & \second{58.5} & 58.7 & \second{68.3} & \second{76.1} & 73.4 & \second{63.9} & \second{71.5} & \second{69.5}\\
    \rowcolor{gray!35}
    DVIS-DAQ & Swin-L & 49.5 & 51.7 & 54.9 & 65.7 & 73.6 & 70.7 & 61.1 & 68.2 & 66.6\\
    \rowcolor{gray!35}
    DVIS-DAQ & VIT-L & \second{53.7} & 58.2 & \second{59.5} & \second{68.3} & \second{76.1} & 73.5 & 62.4 & 70.8 & 68.0\\
    \rowcolor{gray!35}
    DVIS-DAQ\dag & VIT-L & \highest{57.1} & \highest{62.9} & \highest{62.3} & \highest{69.2} & \highest{76.8} & \highest{75.5} & \highest{64.5} & \highest{72.4} & \highest{70.7}\\
\bottomrule[1.5pt]
\end{tabular}\vspace{-3mm}
\end{table*}

\section{Experiment}
\label{sec:exp}

\subsection{Datasets and Metrics}

We conduct experiments on five mainstream video segmentation benchmarks, including OVIS~\cite{qi2022occluded}, Youtube-VIS 2019 \& 2021 \& 2022~\cite{yang2019video}, and VIPSeg~\cite{miao2022large}. We employ AP and AR as evaluation metrics for the VIS datasets following~\cite{yang2019video}. We utilize Video Panoptic Quality (VPQ) and Segmentation and Tracking Quality (STQ) for the VPS datasets as evaluation metrics following~\cite{kim2020video,miao2022large}. More detailed settings can be found in the supplementary file.


\subsection{Main Experiments}
\label{sec:main result}

\noindent
\textbf{Performance on OVIS dataset.}
The OVIS dataset presents more challenging cases, such as occlusion, fast motion, and complex motion trajectories.
We present the quantitative evaluation results in Tab. \ref{tab:ovis&yt19&yt21}. 
When using ResNet-50~\cite{he2016deep} as the backbone, our method achieves an AP of 38.7 and 43.5 with the offline refiner, surpassing all existing methods. 
Compared to the current SOTA DVIS++~\cite{zhang2023dvis++}, our method shows an improvement of 1.5 AP (38.7 vs. 37.2) and 2.3 AP (43.5 vs. 41.2) with the offline refiner.
When employing a larger backbone, our method achieves even more significant improvements. 
With Swin-L~\cite{liu2021swin} as the backbone, our method outperforms DVIS~\cite{zhang2023dvis} by 3.6 AP (49.5 vs. 45.9). When using VIT-L~\cite{dosovitskiy2020image} as the backbone, our method surpasses DVIS++ by 4.1 AP (53.7 vs. 49.6). Notably, our DVIS-DAQ even outperforms DVIS++ with the offline refiner (53.7 vs. 53.4), while DVIS-DAQ with the offline refiner has reached 57.1 AP. 
These quantitative results demonstrate that our DAQ design significantly enhances the capability of query-based video segmentation methods in processing complex scenes with many object emergence and disappearances.


\noindent
\textbf{Performance on Youtube-VIS 2019 \& 2021 datasets.}
The videos in these two datasets are relatively short and feature simple scenes. 
As shown in Tab.~\ref{tab:ovis&yt19&yt21}, when using ResNet-50 as the backbone, our method achieves a 4.0 AP improvement compared to the baseline DVIS on the YTVIS 2019 and 2021 datasets. 
When using VIT-L as the backbone, we achieve comparable performance with DVIS++.
When using the offline refiner, we outperform DVIS++ by 0.9 AP and 0.6 AP on the YTVIS 2019 and 2021 datasets.

\begin{table}[t!]
\centering
\setlength{\tabcolsep}{4.0pt}
\caption{Results on the validation sets of YouTube-VIS 2022 and VIPSeg. AP$^{\rm L}$ refers to the AP on the long video set. VPQ$^{\rm Th}$ and VPQ$^{\rm St}$ refer to the VPQ on the "thing" objects and the "stuff" objects, respectively.}
\label{tab:yt22&vipseg}
\begin{tabular}{l l | c c c | c c c c}
\toprule[1.5pt]
	\multirow{2}{*}{Method} & \multirow{2}{*}{Backbone} & \multicolumn{3}{c|}{YT-VIS 2022} & \multicolumn{4}{c}{VIPSeg}\\
	~ & ~ & AP$^{\rm L}$ & AP$_{\rm 75}^{\rm L}$ & AR$_{\rm 10}^{\rm L}$ & VPQ &  VPQ$^{\rm Th}$ & VPQ$^{\rm St}$ & STQ \\
	\midrule[1pt]
	MinVIS~\cite{huang2022minvis} & R50 & 23.3 & 19.3 & 28.0 & - & - & - & - \\
        TarVIS~\cite{athar2023tarvis} & R50 & - & - & - & 33.5 & 39.2 & 28.5 & \highest{43.1} \\
	DVIS~\cite{zhang2023dvis} & R50 & \second{31.6} & \second{37.0} & \second{36.3} & 39.4 & 38.6 & 40.1 & 36.3 \\
	DVIS++~\cite{zhang2023dvis++} & R50 & \highest{37.2} & \highest{40.7} & \highest{44.6} & \second{41.9} & \second{41.0} & \highest{42.7} & 38.5 \\
	\rowcolor{gray!35}
        DVIS-DAQ & R50 & \second{34.6} & 35.5 & \second{41.1} & \highest{42.1} & \highest{41.7} & \second{42.5} & \second{39.3}\\
        \hline
	MinVIS~\cite{huang2022minvis} & Swin-L & 33.1 & 33.7 & 36.6 & - & - & - & - \\	
        TarVIS~\cite{athar2023tarvis} & Swin-L & - & - & - & 48.0 & \second{58.2} & 39.0 & \highest{52.9} \\
	DVIS~\cite{zhang2023dvis} & Swin-L & \second{39.9} & \second{42.6} & \second{44.9} & 54.7 & 54.8 & \second{54.6} & 47.7 \\ 
	DVIS++~\cite{zhang2023dvis++} & VIT-L & 37.5 & 39.4 & 43.5 & \second{56.0} & 58.0 & 54.3 & 49.8 \\ 
        \rowcolor{gray!35}
        DVIS-DAQ & VIT-L & \highest{42.0} & \highest{43.0} & \highest{48.4} & \highest{57.4} & \highest{60.4} & \highest{54.7} & \second{52.0}\\
\bottomrule[1.5pt]
\end{tabular}\vspace{-3mm}
\end{table}

\noindent
\textbf{Performance on Youtube-VIS 2022 dataset.}
Youtube-VIS 2022 has been expanded with 71 long videos based on the 2021 version. 
Therefore, we only report the AP$^{\rm L}$ for the long videos in Tab.~\ref{tab:yt22&vipseg}. When using ResNet-50 as the backbone, DVIS-DAQ outperforms the baseline DVIS with 3.0 AP$^{\rm L}$. 
Our method outperforms DVIS++ with 4.5 AP$^{\rm L}$ (42.0 vs. 37.5) when using VIT-L as the backbone.
%

\noindent
\textbf{Performance on VIPSeg dataset.}
VIPSeg is a large-scale video panoptic segmentation dataset containing various real-world scenes and more categories. 
Tab.~\ref{tab:yt22&vipseg} presents the performance comparison with SOTA methods. 
Our method achieves new SOTA performance with both ResNet-50 and VIT-L backbone settings. 
Our method achieves a VPQ of 42.1 when using ResNet-50 as the backbone, surpassing the baseline DVIS by 2.7 VPQ. 
With VIT-L as the backbone, our method achieves a VPQ of 57.4, demonstrating a 2.4 VPQ improvement over the previous SOTA DVIS++ in "thing" objects.


\subsection{Ablation Studies and Analysis}
\label{sec:ablation}
We perform ablation experiments on the OVIS dataset using the ResNet-50 backbone and 40K training iterations. We use DVIS~\cite{zhang2023dvis} as the baseline. When aligning the training and testing settings with DVIS++~\cite{zhang2023dvis++}, the baseline achieves an AP of 35.4. 
More detailed settings are provided in the supplementary materials.
Moreover, we also refer the readers for more different architectures using our DAQ in supplementary part.

\begin{table}[!t]
\scriptsize
\setlength{\belowcaptionskip}{0.05cm}
\caption{Ablation studies on our proposed DAQ and other network details.}
\label{tab:all_ab}
\begin{subtable}[t]{0.3\textwidth}
    \setlength{\belowcaptionskip}{0.0cm}
    \centering
    \caption{The main ablation study of the proposed DAQ and EDS. The baseline is the extension of DVIS~\cite{zhang2023dvis}. EDAQ/DDAQ denotes emergence/disappearance DAQ. ES/DS denotes emergence/disappearance simulation.}
    \label{tab:main ablation}
    \setlength{\tabcolsep}{2pt}
    \begin{tabular}[t]{c c c c| c}
    \toprule[1.5pt]
    EDAQ & ES & DDAQ & DS & AP \\
    \midrule[1pt]
    ~ & ~ & ~ & ~ & 35.4 \\
    \checkmark & ~ & ~ & ~ & 33.3 \\
    ~ & \checkmark & ~ & ~ & 32.3 \\
    \checkmark & \checkmark & ~ & ~ & 36.4 \\
    \checkmark & \checkmark & \checkmark & ~ & 36.6 \\
    \checkmark & \checkmark & \checkmark & \checkmark & 37.1 \\
    \bottomrule[1.5pt]
    \end{tabular}
\end{subtable}
\hspace{\fill}
\begin{subtable}[t]{0.2\textwidth}
    \setlength{\belowcaptionskip}{0.0cm}
    \centering
    \caption{Candidates selection for emergence DAQ. K denotes the number of top-scoring candidates. N denotes the number of learnable embeddings.}
    \label{tab:num combination of daq}
    \renewcommand{\arraystretch}{1.4}
    \setlength{\tabcolsep}{4pt}
    \begin{tabular}[t]{c c|c }
    \toprule[1.5pt]
    K & N & $AP $  \\
    \midrule[1pt]
    100 & 1 & 36.4 \\
    50 & 1 & 35.9 \\
    50 & 2 & 36.5 \\
    25 & 4 & 35.6 \\
    \bottomrule[1.5pt]
    \end{tabular}
\end{subtable}
\hspace{\fill}
\begin{subtable}[t]{0.18\textwidth}
    \flushright
    \setlength{\belowcaptionskip}{1.4cm}
    \centering
    \caption{Threshold selection for new emergence simulation.}
    \label{tab:new_sim_thresh}
    \renewcommand{\arraystretch}{1.15}       
    \setlength{\tabcolsep}{2pt}
    \begin{tabular}[t]{c|c}                  
         \toprule[1.5pt]
         Threshold & AP \\
         \midrule[1pt]
         0.01 & 34.2 \\
         0.05 & 34.7 \\
         0.10 & 35.6 \\
         0.20 & 34.5 \\
         0.50 & 33.2 \\
         \bottomrule[1.5pt]
    \end{tabular}
\end{subtable}
\hspace{\fill}
\begin{subtable}[t]{0.25\textwidth}
    \setlength{\belowcaptionskip}{0.0cm}
    \centering
    \caption{Disappearance simulation. \emph{One}, \emph{Continuous}, and \emph{Random} represent simulating disappearance for an object in a solitary frame, all frames, and randomly selected frames within a video clip, respectively.}
    \label{tab:dis_sim_type}
    \renewcommand{\arraystretch}{1.7}       
    \setlength{\tabcolsep}{4pt}
    \begin{tabular}[t]{c|c}
         \toprule[1.5pt]
         Type & AP \\
         \midrule[1pt]
         One & 36.3 \\
         Continuous & 35.8 \\
         Random & 37.1 \\
         \bottomrule[1.5pt]
    \end{tabular}
\end{subtable}

\begin{subtable}[t]{0.48\textwidth}
    \setlength{\belowcaptionskip}{0.3cm}
    \centering
    \caption{Generation of emergence DAQ. \emph{SegQ}/\emph{SegP}/\emph{AppF} denotes the query feature/positional embedding/appearance feature of the segmenter's frame object output.  \emph{Add}/\emph{Cat} denotes adding/concatenating the selected features to the query feature of emergence DAQ. \emph{Pos} denotes using the selected feature as the positional embedding of emergence DAQ.}
    \label{tab:gen_daq}
    \begin{tabular}[t]{c|c c c | c c c}
     \toprule[1.5pt]
     \multirow{2}{*}{Metric} & \multicolumn{3}{c|}{Feature selection} & \multicolumn{3}{c}{Feature usage} \\
     ~ & \emph{SegQ} & \emph{SegP} & \emph{AppF}  & \emph{Add} & \emph{Cat} & \emph{Pos} \\
     \midrule[1pt]
     $AP$ & 34.8 & 36.0 & 36.4 & 36.0 & 36.2 & 36.4 \\
     \bottomrule[1.5pt]
    \end{tabular}
\end{subtable}
\hspace{\fill}
\begin{subtable}[t]{0.48\textwidth}
    \setlength{\belowcaptionskip}{0.0cm}
    \centering
    \caption{Generation of disappearance DAQ. \emph{Learn} and \emph{Initial SegQ} represent initializing the query feature of disappearance DAQ by using a new learnable embedding and the initial queries of segmenter, respectively. \emph{CTQF}/\emph{CTQP} represent using the query feature/momentum-weighted appearance feature of tracked objects as the positional embeddings of disappearance DAQ. }
    \label{tab:dis.DAQ}
    \begin{tabular}[t]{c|c c | c c}
        \toprule[1.5pt]
        \multirow{2}{*}{Metric} & \multicolumn{2}{c|}{Query feature} & \multicolumn{2}{c}{Positional embedding}  \\
        ~ & \emph{Learn} & \emph{Initial SegQ} & \emph{CTQF} & \emph{CTQP} \\
        \midrule[1pt]
        AP & 36.5 & 37.1 & 36.8 & 37.1 \\
        \bottomrule[1.5pt]
    \end{tabular}
\end{subtable}

\end{table}

\noindent
\textbf{Effectiveness of DAQ and EDS.}
Tab.~\ref{tab:main ablation} presents the ablation studies about our proposed components. Using emergence dynamic anchor queries alone leads to a performance degradation of 2.1 AP, attributed to insufficient examples of emergence objects for training. When the emergence simulation strategy is used alone, performance drops by 3.1 AP, indicating that the unreasonable anchor design in the previous method~\cite{zhang2023dvis} makes it difficult for the network to learn to handle object emergence, even with sufficient training cases. When emergence dynamic anchor queries and emergence simulation are used together, the reasonable mechanism and sufficient training cases result in a 1.0 AP improvement compared to the baseline. Disappearing dynamic anchor queries contribute to a performance improvement of 0.2 AP. Introducing the Disappearing Simulation strategy to provide sufficient cases of disappearing objects further increases model performance by 0.5 AP.

\noindent
\textbf{Emergence dynamic anchor queries.} DAQ comprises query feature embedding and positional embedding. Firstly, we explore different strategies for selecting emergency candidates, and the results are presented in Tab.~\ref{tab:num combination of daq}. 
When choosing the top 100 object predictions of the segmenter as candidates, with only one shared learnable embedding, we obtain 36.4 AP. Setting the number of candidates to the top 50 and the number of learnable embeddings to 2 brings a performance gain of 0.1 AP.  To explore potential new objects fully, we choose the top 100 scheme.

In Tab.~\ref{tab:gen_daq}, we investigate how to generate corresponding dynamic anchor queries based on the features of emerging candidates. 
We find that utilizing candidates' appearance features yields the best results, surpassing the use of positional features and object queries with a hybrid of positional and appearance information by 1.6 AP and 0.4 AP, respectively. 
Furthermore, leveraging the features of candidates in the form of query positional embedding results in the best performance, outperforming addition and concatenation operations by 0.4 AP and 0.2 AP, respectively.

\noindent
\textbf{Disappearance dynamic anchor queries.}
In Tab. \ref{tab:dis.DAQ}, we explore how to initialize the query feature embedding and positional embedding of disappearance DAQ. 
We find that initializing query feature of disappearance DAQ from the initial queries in the segmenter (as shown in Fig.~\ref{fig:daq}) yields better AP than using a new shared learnable embedding for all disappearance DAQ (37.1 vs. 36.5). 
For generation of positional embedding of disappearance DAQ, we find utilizing the momentum-weighted appearance feature of tracked objects yield better AP than utilizing the query feature of tracked objects (37.1 vs. 36.8).

\noindent
\textbf{Emergence and disappearance simulation.}
We employ a threshold to filter out low-scoring tracked objects in emergence simulation. 
Tab. \ref{tab:new_sim_thresh} illustrates the results when using different thresholds. A minimal threshold (0.01) results in insufficient new emergence cases, whereas a larger threshold (0.5) disrupts the model's ability to learn continuous tracking. Both result in poor performance (34.2 AP and 33.2 AP). A similar phenomenon is observed in the disappearance simulation, as shown in Tab. \ref{tab:dis_sim_type}. 
The simulation of an object's disappearance in a solitary frame within a video clip fails to generate sufficient disappearance cases. 
Conversely, simulating the disappearance across all frames compromises the model's capability to learn continuous tracking. Consequently, we have determined that a threshold of 0.1 strikes an optimal balance for simulating emergence, and we opt for a stochastic selection of frames in a video clip to simulate disappearance. 
This strategy ensures ample emergence and disappearance cases while preserving the model's tracking capability for continually appearing objects.

\begin{figure}[t]
    \centering
    \includegraphics[width=0.85\linewidth]{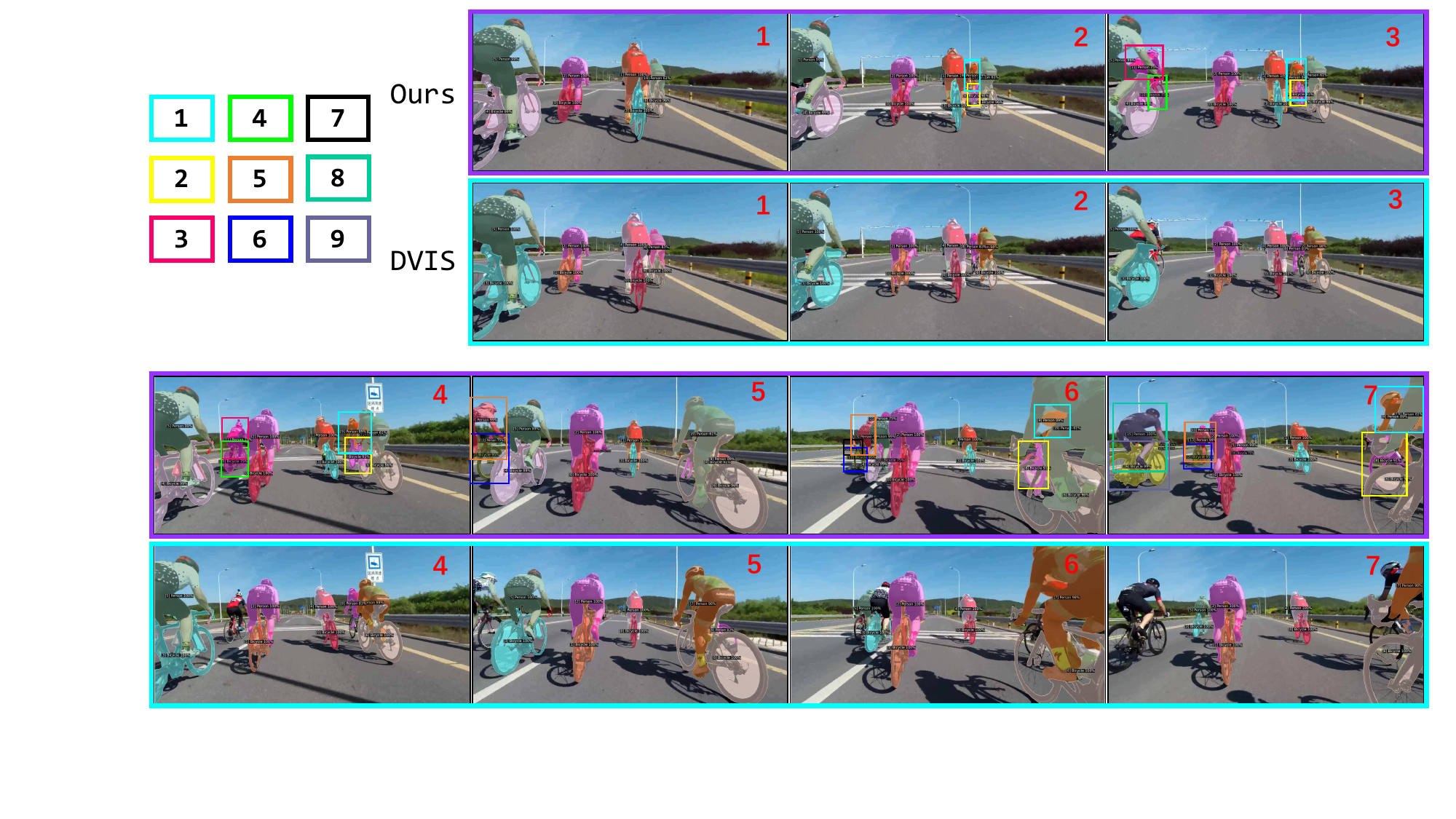}
    \caption{Qualitative results comparison for emergence detection and disappearance filtering. The performance enhancements are highlighted by rectangles in various colors, each color representing an individual video object. It's better to view with zooming in 8$\times$.}
    \label{fig:visual_1}\vspace{-5mm}
\end{figure}

\noindent
\textbf{Qualitative results comparison.}
In Fig. \ref{fig:visual_1}, we present the visual comparison with our baseline. In this challenging scene, there are new-emerging, disappearing, and reappearing video objects. Our method accurately addresses all these scenarios, whereas DVIS suffers from issues including missed new-emerging objects and ID switches. 


\section{Conclusion}
\label{sec:conclusion}

We observed that current SOTA query-based video segmentation methods perform poorly on newly emerging and disappearing objects. 
We have analyzed and identified the significant transition gap hindering the network's learning. 
We address this challenge by proposing dynamic anchor queries for shortening the transition gap. 
We also introduce a query-level simulation strategy for simulating object disappearance and emergence without additional costs.
By integrating our proposed dynamic anchor queries and simulation strategy into DVIS, we obtained DVIS-DAQ, which achieved SOTA performance on mainstream video segmentation benchmarks. 
%
%
Our research will inspire the video community to bridge the gap between current advancements in the standard benchmarks and real-world application requirements.



%
%
\bibliographystyle{splncs04}
\bibliography{main}
\newpage
\appendix
\setcounter{figure}{0}
\renewcommand{\thefigure}{A\arabic{figure}}
\setcounter{table}{0}
\renewcommand{\thetable}{A\arabic{table}}

\newpage

\section{Appendix}

\appendix

\noindent
\textbf{Overview.} In this section, we first introduce more implementation details and training settings of our model. 
To further substantiate the generalizability and effectiveness of our DAQ design, we integrate the DAQ into an alternative video segmentation framework, GenVIS~\cite{heo2023generalized}, and conduct experiments on the OVIS~\cite{qi2022occluded} dataset.
Additionally, we provide more ablation studies to verify the effectiveness of the detailed components of the DAQ design. 
Lastly, we present a collection of qualitative results and supplementary video files to demonstrate the full performance of our method in challenging scenes.

\subsection{More Implementation Details}
\label{sec:exp_details_supp}

\noindent
\textbf{Spatio-temporal padding.}
The tracker incorporating with DAQ outputs $N_{trc}$ video objects for a video with $T$ frames, denoted as $ \{\{Q^i\}^{T_i}_{t_i}\}^{N_{trc}}_{i=1}$, where the temporal dimension lengths ($T_i - t_i + 1$) of these $N_{trc}$ video objects may vary. Therefore, before feeding into the temporal refiner, we pad each video object sequence with the momentum-weighted appearance feature at the time steps $t \in \{1,\ldots,t_i-1, T_i+1,\ldots, T\}$, where the video object does not appear. The padded results are denoted as $\{\{Q^i\}^T_1\}^{N_{trc}}_{i=1}$. In addition to temporal padding, we also perform padding in the spatial dimension. We first use the Hungarian algorithm to naively extract $N$ video object feature sequences $\{ \{Q_{naive}^i\}^{T}_{1} \}^N_{i=1}$ from the object queries $Q_{Seg} \in R^{ N \times C}$ of all frames, as MinVIS \cite{huang2022minvis} does. Then, we select the top $N-N_{trc}$ video objects based on classification scores to pad $\{\{Q^i\}^T_1\}^{N_{trc}}_{i=1}$ in the spatial dimension to obtain $\{\{Q^i\}^T_1\}^N_{i=1}$.

\noindent
\textbf{Training details.}
We employ the AdamW optimizer with an initial learning rate of 1e-4 and a weight decay of 5e-2 for training. For the VIS task, we use the COCO joint training setting to train the model. For the VPS task, no additional datasets are employed. When training in offline mode, we freeze all modules except the temporal refiner and initialize these modules with parameters trained in online mode. We set the lengths of training video clips as 5 and 15 for online and offline modes, respectively. For all datasets, we train for 160K iterations and apply learning rate decay at 112K iterations.

\subsection{Additional Experiments}
\label{sec:addition_exp_supp}
\noindent\textbf{Generalization ability of DAQ.} To further substantiate the efficacy of the Dynamic Attention Query (DAQ), we integrated it into an alternative video segmentation framework, GenVIS~\cite{heo2023generalized}, culminating in the development of an augmented architecture dubbed GenVIS-DAQ. We employed ResNet-50 as the backbone and conducted experiments on the more challenging OVIS dataset. For a fair comparison, we used GenVIS without the instance prototype memory (IPM) as our baseline and adhered to the original training strategies utilized by GenVIS. The results are reported in Tab.~\ref{tab:genvis-daq}. By incorporating DAQ, we achieved an improvement of 1.7 AP for GenVIS. This demonstrates that using DAQ to manage the emergence and disappearance of objects in videos can enhance the capability of query-based video segmentation methods to handle complex scenes.

\noindent\textbf{Performance improvements on emergence and disappearance.} We conduct separate evaluations for newly emerged and disappeared objects on the OVIS and VIPSeg datasets, with the results presented in Tab~.\ref{tab:ed_performance}. For newly emerged and disappeared objects, our method outperforms the baseline DVIS~\cite{zhang2023dvis} by 7.3 VPQ and 6.8 AP on VIPSeg and OVIS, respectively, indicating the effectiveness of our method for this problem.

\noindent\textbf{Feature transition gap.} We claimed that our DAQ mechanism enhances the management of objects' emergence and disappearance by shortening the feature transition gap between the anchor query and the target. We define the transition gap as the feature distance from the anchor query to the target. In DVIS~\cite{zhang2023dvis}, the anchor query for a newly emerged object is a background query, and the target is the feature representation of the object. In contrast, the anchor query for a disappeared object is the feature representation of the object before it disappeared, and the target is a background feature.  In our method, the dynamic anchor query is a mixture of background and foreground for both situations, where learnable embeddings and candidates provide the background and foreground features, respectively. Our DAQs offer information on potential future states of targets.  Therefore, our method results in a smaller transition gap to the target. As shown in Tab.~\ref{tab:gap}, we calculate the transition gap for DVIS and our method, where our DAQ significantly shortens the transition gap.

\begin{table}
\centering
\setlength{\belowcaptionskip}{0cm}
\setlength{\tabcolsep}{2.0pt}
\caption{Results on the validation sets of OVIS. }
\label{tab:genvis-daq}
\setlength{\tabcolsep}{3pt}
\begin{tabular}{l | c c c c c }
\toprule[1.5pt]
\multirow{2}{*}{Method} & \multicolumn{5}{c}{OVIS}\\
~ & AP &  AP$_{\rm 50}$ & AP$_{\rm 75}$ &  AR$_{\rm 1}$ & AR$_{\rm 10}$ \\
\midrule[1pt]
Baseline & 34.8 & 59.0 & 35.4 & 17.0 & 39.0 \\
GenVIS~\cite{heo2023generalized} + DAQ & 36.5 \textcolor[rgb]{1, 0, 0}{(+1.7)} & 64.5 \textcolor[rgb]{1, 0, 0}{(+5.5)} & 35.6 \textcolor[rgb]{1, 0, 0}{(+0.2)} & 15.4 \textcolor[rgb]{0.66,0.66,0.66}{(-1.6)} & 42.5 \textcolor[rgb]{1, 0, 0}{(+3.5)} \\
\bottomrule[1.5pt]
\end{tabular}
\end{table}

\begin{table}
    \centering
    \setlength{\belowcaptionskip}{-0.2cm}
    \caption{Performance improvements. VPQ$^{all}$ and AP$^{all}$ represent performance on all objects, while VPQ$^{ed}$ and AP$^{ed}$ for emerging and disappearing objects.}
    \label{tab:ed_performance}
    \setlength{\tabcolsep}{19pt}
    \begin{tabular}[t]{c|c c| c c}
        \toprule[1.5pt]
             \multirow{2}{*}{Method} & \multicolumn{2}{c|}{VIPSeg} & \multicolumn{2}{c}{OVIS} \\
             ~ & VPQ$^{all}$ & VPQ$^{ed}$ & AP$^{all}$ & AP$^{ed}$ \\
             \midrule[1pt]
             DVIS~\cite{zhang2023dvis} & 38.7 & 29.0 & 26.9 & 19.1 \\
             DVIS+DAQ & 42.1 & 36.3 & 30.2 & 25.9\\
             \midrule[1pt]
             Improvement & \textcolor[rgb]{0, 0, 1}{+3.4} & \textcolor[rgb]{1, 0, 0}{+7.3} & \textcolor[rgb]{0, 0, 1}{+3.3} & \textcolor[rgb]{1, 0, 0}{+6.8}\\
        \bottomrule[1.5pt]
    \end{tabular}
\end{table}

\begin{table}
    \centering
    \setlength{\belowcaptionskip}{0cm}
    \caption{Feature transition gap. CS denotes Cosine Similarity. NED denotes Normalized Euclidean Distance. Symbols $\uparrow$ and $\downarrow$ denote ``larger is better'' and ``smaller is better'', respectively.}
    \label{tab:gap}
    \setlength{\tabcolsep}{34pt}
   \begin{tabular}{c | c c}
        \toprule[1.5pt]
             Methods & CS$\uparrow$ & NED$\downarrow$  \\
             \midrule[1pt]
             DVIS~\cite{zhang2023dvis} & $0.08\pm0.06$ & $0.68\pm0.02$ \\
             DVIS+DAQ & $0.32\pm0.08$ & $0.59\pm0.04$ \\
        \bottomrule[1.5pt]
    \end{tabular}
\end{table}

\subsection{More Ablation Studies}

\noindent
\textbf{Softmax in the disappearance tracker.}
As depicted in Tab.~\ref{tab:tracker2}, compared to the standard cross-attention, which applies SoftMax on the key dimension, we observe that applying SoftMax on the query dimension yields better performance. This is because employing SoftMax on the query dimension prevents different disappearance DAQ from merging the features of the same candidate into the same target query.

\noindent
\textbf{Spatio-temporal padding.} As shown in Tab.~\ref{tab:padding_supp}, before integrating online results into the offline module, it is imperative to apply both temporal and spatial padding. This preparatory step guarantees that the input is appropriately aligned in both time and space to fulfill the requirement of the offline module. Opting to pad with momentum-weighted appearance features rather than zero or learnable padding affords the advantage of tailoring feature information to each individual video object. This tailored padding strategy has been instrumental in achieving a notable improvement in AP (42.7 vs. 38.4). Additionally, performing spatial padding subsequent to temporal padding can further enhance the AP (43.5 vs. 42.7).

\noindent
\textbf{Computation cost analysis.} The computational cost of DVIS-DAQ components was measured by evaluating the parameters, FLOPs, and inference time of the segmenter and tracker with DAQ. Tab. As shown in Tab.~\ref{tab:cost}, the additional tracker \#2 only accounts for 1.6\% of the total computational cost.

\begin{table}[t!]
\scriptsize
\setlength{\belowcaptionskip}{0.05cm}
\caption{More ablation studies on our proposed DAQ and other network details.}
\label{tab:more_ab}
\begin{subtable}[t]{0.4\textwidth}
    \setlength{\belowcaptionskip}{0.cm}
    \centering
    \caption{The ablation study of the softmax dimension of cross-attention in disappearance tracker.}
    \label{tab:tracker2}
    \renewcommand{\arraystretch}{1.4}
    \setlength{\tabcolsep}{4pt}
    \begin{tabular}[t]{c|c}
         \toprule[1.5pt]
         Dimension & AP \\
         \midrule[1pt]
         K-dim & 36.6 \\
         Q-dim & 37.1 \\
         \bottomrule[1.5pt]
    \end{tabular}
\end{subtable}
\hspace{\fill}
\begin{subtable}[t]{0.6\textwidth}
    \setlength{\belowcaptionskip}{0.cm}
    \centering
    \caption{Computational cost. Segmenter is implemented by Mask2Former~\cite{cheng2021mask2former} with ResNet-50 as backbone. Input frames are resized to 480p.}
    \label{tab:cost}
    \renewcommand{\arraystretch}{1.15}
    \setlength{\tabcolsep}{4pt}
    \begin{tabular}{c|c c c}
        \toprule[1.5pt]
             Component & Params (M) & FLOPs (G) & Time (ms) \\
             \midrule[1pt]
             Segmenter & 44.15 & 225.14 & 50.01 \\
             Tracker \#1 & 9.94 & 3.78 & \multirow{2}{*}{10.32} \\
             Tracker \#2 & 8.68 & 3.67 & ~ \\
        \bottomrule[1.5pt]
    \end{tabular}
\end{subtable}

\begin{subtable}[t]{1\textwidth}
    \centering
    \setlength{\belowcaptionskip}{0.0cm}
    \caption{Padding the online output queries to offline inputs. For temporal padding, \emph{None} denotes using attention masks in the offline module instead of padding; 
    \emph{Zero}, \emph{Learn}, and \emph{AppF} denote padding with zero value, a learnable embedding, and the momentum-weighted appearance feature, respectively. For spatial padding, \emph{No} denotes not to pad, and \emph{Yes} denotes padding with naively associated video objects.}
    \label{tab:padding_supp}    
    \setlength{\tabcolsep}{15pt}
    \begin{tabular}[t]{c|c c c c | c c}
         \toprule[1.5pt]
         \multirow{2}{*}{Metric} & \multicolumn{4}{c|}{Temporal} & \multicolumn{2}{c}{Spatial} \\
         ~ & None & Zero & Learn & AppF & No & Yes \\
         \midrule[1pt]
         AP & 38.4 & 39.0 & 39.1 & 42.7 & 42.7 & 43.5  \\
         \bottomrule[1.5pt]
    \end{tabular}
\end{subtable}
\end{table}

\subsection{More Qualitative Results}

\begin{figure}
    \centering
    \includegraphics[width=0.85\linewidth]{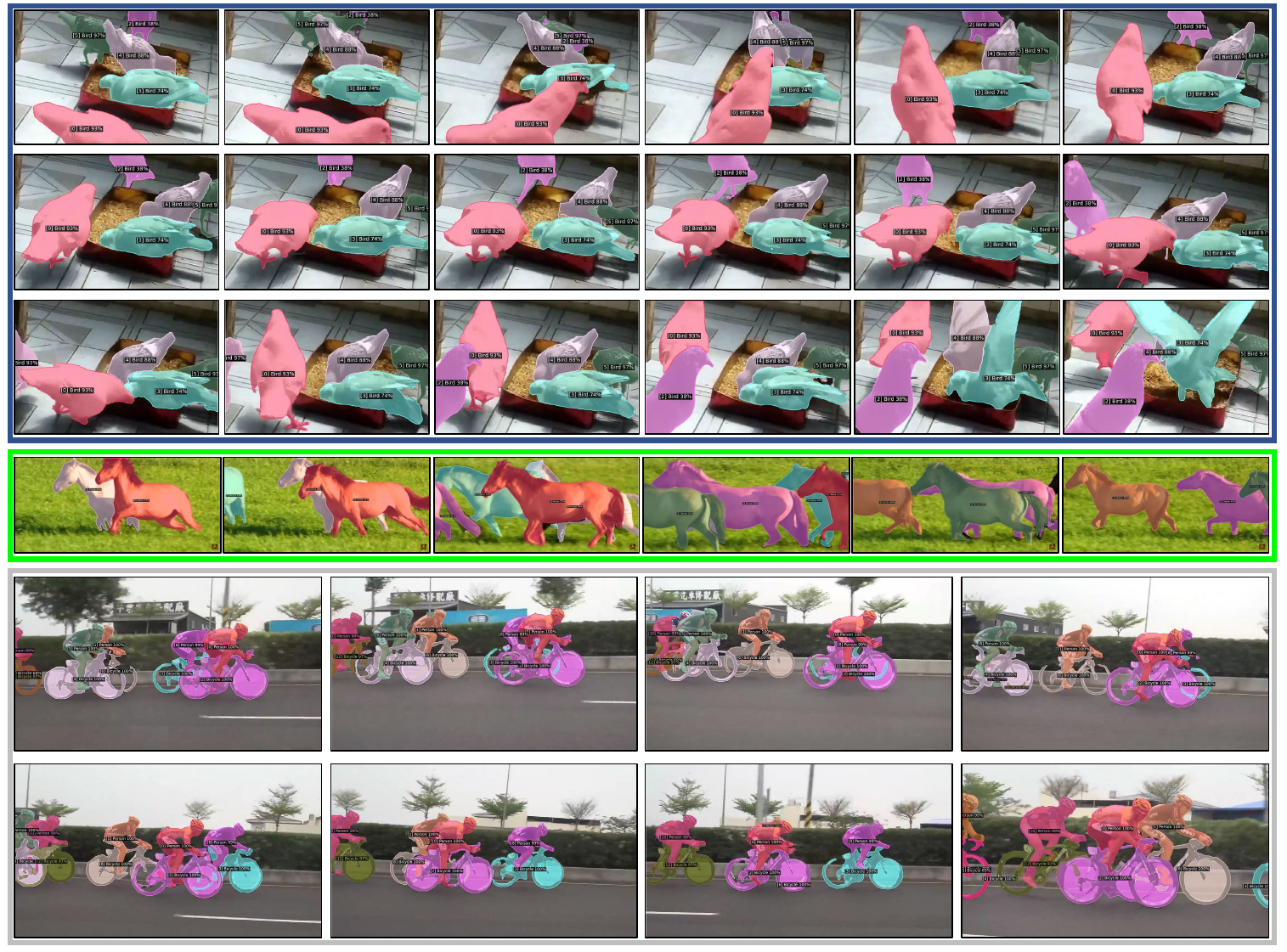}
    \caption{Qualitative results. It's better to view with zooming in 8$\times$.}
    \label{fig:more_vis2}
\end{figure}

Fig.~\ref{fig:more_vis2} presents qualitative results in complex scenes. To fully demonstrate the performance of our method in challenging scenes such as fast motion and occlusions, we also provide supplementary videos for reference at project page.

\end{document}